\documentclass[letterpaper, 10 pt, conference]{ieeeconf}  

\IEEEoverridecommandlockouts                              

\usepackage{graphicx} 

\title{\LARGE 
Towards Persistent Storage and Retrieval of\\ Domain Models using Graph Database Technology
}

\author{Nico Hochgeschwender,$^{1,}$$^{2}$ Holger Voos,$^{2}$ and Gerhard K. Kraetzschmar$^{1}$
\thanks{*Nico Hochgeschwender is recipient of a PhD scholarship from the Graduate Institute of the Bonn-Rhein-Sieg University, which he gratefully acknowledges.}
\thanks{$^{1}$Nico Hochgeschwender, Artem Vinokurov, and Gerhard K. Kraetzschmar are with the Department of Computer Science, Bonn-Rhein-Sieg University, Sankt Augustin, Germany. \texttt{nico.hochgeschwender@h-brs.de}. }%
\thanks{$^{2}$Nico Hochgeschwender and Holger Voos are with the University of Luxembourg, Luxembourg.
}%
}

\begin{document}

\maketitle
\thispagestyle{empty}
\pagestyle{empty}

\begin{figure}[h]
\centering
\includegraphics[width=2.5cm]{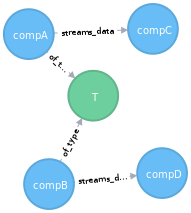}
\caption{Simplified example of a perception graph repository}
\label{fig:graph}
\end{figure}

Robots are expected to perform a wide range of challenging tasks in dynamic environments. To do so robots need to extract knowledge about the world from the data perceived through the sensors of the robot. However, as Crowley \emph{et al.}~\cite{crowley} already pointed out the huge variations in operating conditions such as environmental changes (e.g. illumination and occlusion) and failures (e.g. sensor failures) makes the design, development and deployment of robot perception architectures a challenging and knowledge-intensive exercise. To easily configure, modify and validate robot perception architectures we proposed in our previous work the domain-specific language RPSL (Robot Perception Specification Language)~\cite{RPSL}. The RPSL is a Ruby-based internal DSL which enables to specify individual configurations (called \emph{perception graphs}) of RPAs meeting the requirements of different context conditions. To this end RPSL enables to specify two crucial elements of perception systems, namely perception graphs and data types. In RPSL a perception graph is a composition of components distinguished in sensor and processing components in 
the form of a directed acyclic graph (DAG). As the RPSL is an internal DSL the resulting \emph{domain models} of concrete perception graphs are Ruby code itself. Up to now we stored these domain models in a self-made repository implemented with standard Ruby collections such as hashes, sets, and ranges. We used the repository mainly to facilitate the selection of 
perception graphs during runtime~\cite{etfa}. However, as the repository is steadily growing e.g., new perception graphs 
are added or existing perception graphs are modified a more structured approach to store and retrieve domain models is required.   
A common approach to organise data in a structured way is to employ some sort of database management system (DBMS). In the context of this work we decided to employ a graph database (GD)\footnote{We make use of the \texttt{http://neo4j.com/} graph database.} to store and retrieve RPSL domain models. A GD stores information
in the form of a graph where nodes represent entitites and edges among nodes represent relations. As RPSL domain models are already structured as a graph the decision to choose a GD as DBMS is feasible. Beyond that, a GD enables to express so called \emph{semantic queries} in order to retrieve and infer information based on the relations among the domain models. By doing so we can link the \emph{domain models} as exemplified in the following simplified RPSL repository. Let us assume we have two perception graphs, each composed of two components (see Fig..~\ref{fig:graph}). In both perception graphs the first component produces some output conforming to a type $T$. Here, $T$ is a separated RPSL domain model dealing with data type information. In our repository we link both domain models as seen in the Figure.  
This enables use to perform sophisticated queries incorporating both domain models such as 
retrieve all the components producing type $T$, or 
get all the perception graphs where at least one component produces type $T$.    
As we use the \texttt{neo4j} GD we can utilize the \emph{Cypher DSL} a dedicated SQL-like query language for
GDs. For instance, to retrieve all the components producing type $T$ the following Cypher statement is sufficient:
\begin{verbatim}
match (n:Component), (m:Type) 
where (n)-[:of_type]->(m) return n;
\end{verbatim}
Up to now this language and the GD approach in general is promising. Roughly speaking, it enables to
perform two general types of queries, namely meta-level queries to retrieve general information (e.g. number of components, graphs etc.) about the repository and semantic queries incorporating several domain models.









\begin{thebibliography}{99}
\bibitem{crowley} Crowley, James L. and Hall, Daniela and Emonet, Remi. \emph{Autonomic Computer Vision Systems}. The 5th International Conference on Computer Vision Systems. 2007.
\bibitem{RPSL} Hochgeschwender, Nico and Schneider, Sven and Voos, Holger and Kraetzschmar, Gerhard. \emph{Declarative Specification of Robot Perception Architectures}. Simulation, Modeling, and Programming for Autonomous Robots. Volume 8810 of the series Lecture Notes in Computer Science. 2014.
\bibitem{etfa} Hochgeschwender, Nico and Olivares-Mendez, Miguel A. and Voos, Holger and Kraetzschmar, Gerhard. \emph{Context-based Selection and Execution of Robot Perception Graphs}. IEEE International Conference on Emerging Technologies and Factory Automation (ETFA). 2015. 
\end{thebibliography}
\end{document}